\pdfoutput=1

\documentclass[11pt]{article}

\usepackage{acl}

\usepackage{times}
\usepackage{latexsym}

\usepackage[normalem]{ulem} 
\usepackage{xcolor}

\usepackage[T1]{fontenc}

\usepackage[utf8]{inputenc}

\usepackage{microtype}

\usepackage{inconsolata}

\usepackage[english]{babel}  
\addto\extrasenglish{

}

\usepackage{svg}
\usepackage{tabularx}
\usepackage{multirow}

\usepackage{amsmath}
\usepackage[noabbrev,capitalize]{cleveref}
\usepackage{caption}
\usepackage{subcaption}
\usepackage{enumitem}

\newcolumntype{C}{>{\centering\arraybackslash}X}

\usepackage{pdfpages}

\title{End-to-end Automatic Speech Recognition and Speech Translation:\\Integration of Speech Foundational Models and LLMs}

\author{Nam Luu \and Ond\v{r}ej Bojar \\
  Charles University \\
  Faculty of Mathematics and Physics \\
  Institute of Formal and Applied Linguistics \\
  \texttt{\{luu,bojar\}@ufal.mff.cuni.cz}} 

\begin{document}
\maketitle
\begin{abstract}
Speech Translation (ST) is a machine translation task that involves converting speech signals from one language to the corresponding text in another language;
this task has two different approaches, namely the traditional \emph{cascade} and the more recent \emph{end-to-end}. This paper explores a combined end-to-end architecture of pre-trained speech encoders and Large Language Models (LLMs) for performing both Automatic Speech Recognition (ASR) and ST simultaneously. Experiments with the English-to-German language pair show that our best model not only can achieve better translation results than SeamlessM4T \citep{communication2023seamlessm4tmassivelymultilingual}, a large foundational end-to-end, multi-modal translation model, but can also match the performance of a cascaded system with Whisper \citep{radford2022robustspeechrecognitionlargescale} and NLLB \citep{nllbteam2022languageleftbehindscaling}, with up to a score gain of 8\% in $\text{COMET}^{\text{DA}}_{22}$ metric.
\end{abstract}

\section{Introduction}
\label{sec:intro}


End-to-end Speech Translation is a growing research direction that aims to ignore the intermediate ASR step to directly translate the audio input into its corresponding text in another language. This approach simplifies the overall architecture, which has been shown to match the performance of the cascaded counterpart \citep{bérard2018endtoendautomaticspeechtranslation, liu2019endtoendspeechtranslationknowledge, gaido2020endtoendspeechtranslationknowledgedistillation}.

Large Language Models (LLMs) have demonstrated their emergent capabilities on a large number of complex natural language tasks, including machine translation \citep{minaee2024largelanguagemodelssurvey, zhang2024instructiontuninglargelanguage, zhao2023surveylargelanguagemodels, naveed2024comprehensiveoverviewlargelanguage}. With the ever-improving potential of LLMs, researchers have been trying to integrate different components used for other modalities, in order to extend their abilities to go beyond text-only tasks \citep{li2023blip2bootstrappinglanguageimagepretraining, gao2023llamaadapterv2parameterefficientvisual, liu2023visualinstructiontuning, li2023promptinglargelanguagemodels, zhang2023speechgptempoweringlargelanguage}.

Motivated by recent contributions in speech representation learning and LLMs, we aim to investigate an end-to-end architecture that simultaneously performs both ASR and ST. This architecture combines the high-quality audio representation from the pre-trained acoustic models with the excellent performance of LLMs to serve as an end-to-end speech translation system, 
while still having the ability to transcribe from the audio signal. Our proposed model, after being fine-tuned with the Quantized Low-Rank Adaptation (QLoRA; \citealp{dettmers2023qloraefficientfinetuningquantized}) technique, achieves a robust translation performance, comparable to a cascaded system, which is still a state-of-the-art approach for this task. 

The paper is structured as follows:
\begin{itemize}[noitemsep,nolistsep]
    \item \cref{sec:methods} describes the details of the pipeline, along with the dataset used for training and evaluation.
    \item \cref{sec:eval} provides the ASR and ST evaluation results of the model in different public test sets, and compares them to some baselines from out-of-the-box models.
    \item \cref{sec:future_work} 
    proposes possible directions to improve the architecture.
\end{itemize}

\section{Methods and Dataset}
\label{sec:methods}

\subsection{Architecture}
\begin{figure}[htbp]
  \centering
  \includegraphics[width=\linewidth]{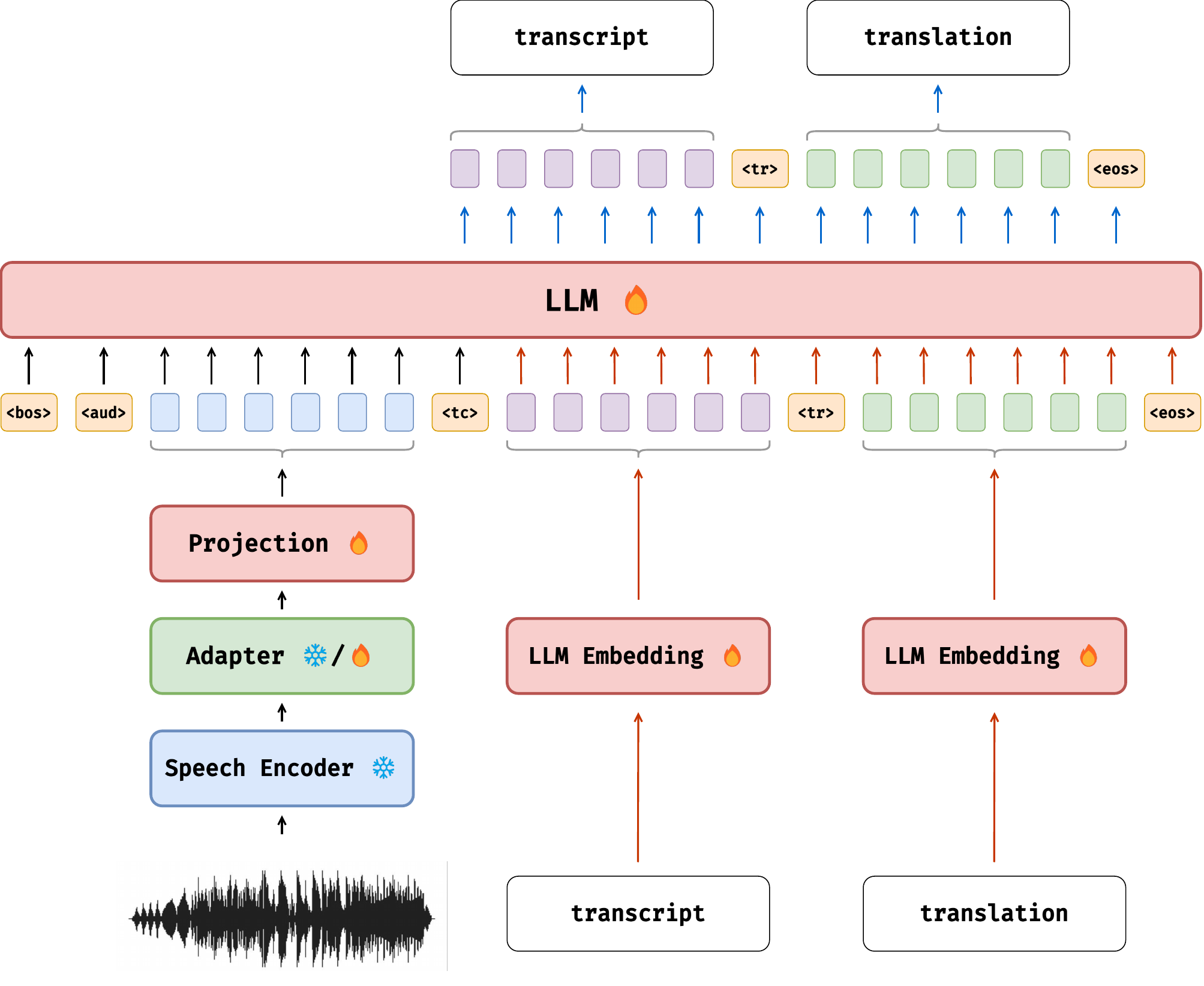}
  \caption{
  The overall architecture includes a frozen speech encoder component, an adapter, and a fine-tuned LLM. The adapter can be frozen or trainable depending on the adapter type. \textcolor{red}{\textbf{Red}} arrows denote the usage of tokens during training, and \textcolor{blue}{\textbf{blue}} arrows indicate tokens generated during inference; while \textbf{black} arrows represent the prompt fed to the LLM.}
  \label{fig:arch}
\end{figure}

The overall architecture is illustrated in \cref{fig:arch}. For each training sample, given the speech signal, its corresponding transcript, and the translated text, the speech hidden features are obtained using a speech encoder, including HuBERT \citep{hsu2021hubertselfsupervisedspeechrepresentation} and Whisper encoder \citep{radford2022robustspeechrecognitionlargescale}. 

Next, the speech features are fed to a Projection layer, in order to convert the feature dimension to match the LLM's embedding dimension. The resulting speech embeddings are subsequently given to the LLM as the prompt for it to generate the corresponding transcription and the translated text simultaneously. The LLM is then fine-tuned in the next-token-prediction fashion.

\subsection{Speech Encoder}
We adopted HuBERT \citep{hsu2021hubertselfsupervisedspeechrepresentation} and Whisper \citep{radford2022robustspeechrecognitionlargescale} as the speech encoders, utilizing their capability of extracting high-quality representation from audio data. We used the \texttt{hubert-large-ls960-ft}
variation, which was trained on 60,000 hours of data from the LibriLight \citep{Kahn_2020} corpus, then fine-tuned on 960 hours of data from the LibriSpeech \citep{librispeech7178964} corpus. For Whisper-based models, we only used the encoder part of the pre-trained \texttt{whisper-large-v3-turbo}
to extract the audio hidden features.

\subsection{Length Adapter}
Because the length of the speech feature sequence can be longer than the supported length of the LLM, it is more favorable to shorten it beforehand.

For HuBERT-based models, we followed the work of \citet{gaido2021ctcbasedcompressiondirectspeech}, and compressed the feature sequence by taking an average of vectors whose repeated labels were obtained from the followed Connectionist Temporal Classification (CTC) layer. \citet{wu2023decoderonlyarchitecturespeechtotextlarge} illustrated that speech feature sequence compression with CTC gave better results than the traditional collapsing approach with convolution layers in the speech translation task. Hence, in our pipeline, from the obtained labels predicted by CTC, we merged the vectors with repeating labels by averaging their corresponding values.

\begin{figure}[htbp]
  \centering
  \begin{subfigure}[b]{0.62\linewidth}
    \centering
    \includegraphics[width=\linewidth]{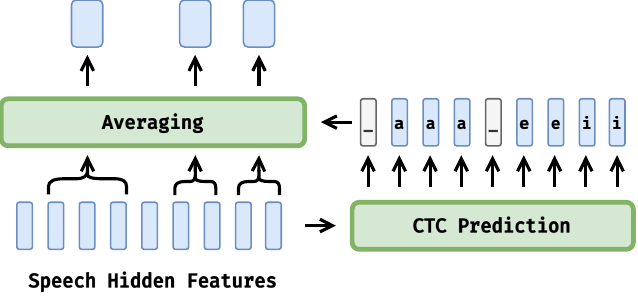}
    \caption{CTC collapse}
    \label{fig:arch_ctc}
  \end{subfigure}
  \hfill
  \begin{subfigure}[b]{0.305\linewidth}
    \centering
    \includegraphics[width=\linewidth]{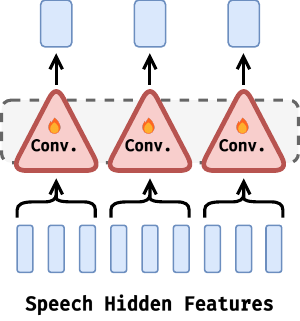}
    \caption{Convolution}
    \label{fig:arch_conv}
  \end{subfigure}
\caption{Details of different adapters}
\label{fig:arch_adapter}
\end{figure}

While for Whisper-based models, a convolution-based downsampling layer with a kernel size of 5 and a stride of 5 is used to reduce the length of the speech feature sequence. The details of both length adapters are illustrated in \cref{fig:arch_adapter}.

\subsection{Projection Layer}
For the Projection layer, we used only one simple feed-forward layer to map from the encoder's hidden size to the corresponding LLM's hidden size. This layer ensures the resulting speech representation is well integrated into the LLM's embedding space, giving it enough information for the downstream task.

\subsection{LLMs}
We experimented with four different pre-trained LLMs available on HuggingFace, namely \emph{Gemma 7B} (\texttt{gemma-7b}),
\emph{Gemma 2 9B} (\texttt{gemma-2-9b}),
\emph{Llama 2 7B} (\texttt{Llama-2-7b-hf}),
and \emph{Mistral 7B v0.1} (\texttt{Mistral-7B-v0.1}).
Details about each variation are described in \cref{tab:detail}.

\begin{table}[h]
\small
\centering
{\renewcommand{\arraystretch}{1.1}
\begin{tabularx}{0.49\textwidth}{p{0.1\textwidth}lC}
\hline
\multicolumn{1}{c}{\textbf{Encoder}}  & \multicolumn{1}{c}{\textbf{Decoder}} & \textbf{Adapter} \\ \hline
\multirow{4}{*}{HuBERT}  & Gemma 7B                              & \multirow{4}{\linewidth}{CTC collapse}                                                           \\ 
                                        & Gemma 2 9B                            &                                                                        \\ 
                                        & Llama 2 7B                            &                                                                        \\ 
                                        & Mistral 7B v0.1                       &                                                                     \\ \hline
\multirow{4}{\linewidth}{Whisper enc.} & Gemma 7B                              & \multirow{4}{\linewidth}{5x5 Convolution}                                                        \\ 
                                        & Gemma 2 9B                            &                                                                    \\
                                        & Llama 2 7B                            &                                                                        \\
                                        & Mistral 7B v0.1                       &                                                                      \\ \hline
\end{tabularx}}
\caption{Details of each model, with its corresponding Encoder and Decoder components}
\label{tab:detail}
\end{table}

\subsection{Dataset}
All models were trained using the MuST-C dataset \citep{CATTONI2021101155},
a large multilingual corpus built from English TED Talks, which contains the audio data, the English transcription of such audio, with its translation in multiple languages. In specific, we used the English-to-German subset from version 1.0 of the dataset, with approximately 400 hours of audio data.

For evaluation, MuST-C also provides two public test sets, both named \texttt{tst-COMMON} in version 2.0 and 3.0. We also used the test sets from the Offline Track of IWSLT'21
and '22.
In addition, to evaluate ASR performance, we used two test sets from the LibriSpeech \citep{7178964} dataset, namely \texttt{test-clean} and \texttt{test-other}, both of which are the standard datasets for this task. As all models can perform both ASR and ST simultaneously, evaluation results for both tasks are described in \cref{subsec:asr,subsec:st}, respectively.

\section{Evaluation}
\label{sec:eval}

\subsection{Metrics and Tools}
For the Offline Speech Translation task, we evaluated all models using standard metrics, namely \textsc{BLEU} \citep{bleu}, $\text{COMET}^{\text{DA}}_{22}$ \citep{rei-etal-2022-comet},\footnote{\url{https://huggingface.co/Unbabel/wmt22-comet-da}} and $\text{COMET}^{\text{KIWI-DA}}_{22}$ \citep{rei2022cometkiwiistunbabel2022submission}.\footnote{\url{https://huggingface.co/Unbabel/wmt22-cometkiwi-da}} For the Automatic Speech Recognition Task, we used \textsc{WER}, the standard metric for speech recognition.


For the evaluation purpose, we used the \texttt{SLTev} \citep{ansari-etal-2021-sltev} library,
because it supports both MT and ASR evaluation in one package, using \texttt{sacreBLEU} \citep{post2018clarityreportingbleuscores} to calculate \textsc{BLEU} score. However, since \texttt{SLTev} does not report any \textsc{COMET}-family metrics, we had to change the structure of the sentence with \texttt{mwerSegmenter},\footnote{\url{https://www-i6.informatik.rwth-aachen.de/web/Software/mwerSegmenter.tar.gz}} 
to automatically resegment the models' output according to the reference, 
before evaluating with the \texttt{unbabel-comet}
package. The evaluation was done using \texttt{python-3.11.5}, \texttt{SLTev-1.2.3}, and \texttt{unbabel-comet-2.2.2}.

We compared our architecture with two out-of-the-box baselines: a cascaded pipeline of Whisper
(\texttt{whisper-large-v3-turbo}; \citealp{radford2022robustspeechrecognitionlargescale}) producing the transcript and NLLB
(\texttt{nllb-200-3.3B}; \citealp{nllbteam2022languageleftbehindscaling}) translating the transcript, along with SeamlessM4T
(\texttt{seamless-m4t-v2-large}; \citealp{communication2023seamlessm4tmassivelymultilingual}) - an end-to-end, multi-modal translation model.

\subsection{ASR Results}
\label{subsec:asr}

\cref{tab:asr} details the ASR evaluation results against the four test sets. We reported the \textsc{WER} score after applying the ``LPW'' pre-processing strategy available in \texttt{SLTev}, which first lowercased every character, removed all punctuation, then used the built-in \texttt{mwerSegmenter} tool to resegment the output transcripts. Due to some bugs when processing the IWSLT'21 test set (tst2021), \texttt{mwerSegmenter} failed to run during evaluation, hence we could not obtain the results. It can be seen that models with Gemma 2 9B as the decoder have the best result among the four LLMs, albeit still lagging behind the performance of Whisper.

\begin{table*}[h!]
\small
\centering
{\renewcommand{\arraystretch}{1.1}
\begin{tabularx}{\linewidth}{p{0.25\linewidth}CCCCC}
\hline
\multicolumn{1}{c}{\multirow{2}{*}{\textbf{Model}}} & \multicolumn{2}{c}{\textbf{MuST-C}} & \multicolumn{1}{c}{\textbf{IWSLT}} & \multicolumn{2}{c}{\textbf{LibriSpeech}}  \\
\text{} & \textbf{tst-COMMON v2} & \textbf{tst-COMMON v3} & \textbf{tst2022} &  \textbf{test-clean} & \textbf{test-other} \\ \hline \hline
\multicolumn{1}{l}{Whisper}                      & $\mathbf{6.7\%}$       & $\mathbf{7.7\%}$               & $\mathbf{11.8\%}$    & $\mathbf{4.1\%}$                & $\mathbf{7.2\%}$           \\ \hline \hline
HuBERT + Gemma 2 9B                              & $11.1$\%               & $12.5$\%           
                & $21.9$\%         & $8.4$\%                & $13.1$\%          \\
HuBERT + Gemma 7B                                & $12.9$\%               & $14.5$\%                           & $30.7$\%         & $11.7$\%               & $17.4$\%          \\
HuBERT + Llama 2 7B                              & $11.1$\%               & $12.6$\%                           & $22.9$\%         & $8.7$\%                & $13.2$\%          \\
HuBERT + Mistral 7B v0.1                         & $11.1$\%               & $12.4$\%                           & $22.9$\%         & $8.5$\%                & $13.3$\%          \\ \hline
Whisper enc. + Gemma 2 9B                        & $8.2$\%                & $8.1$\%                            & $22.6$\%         & $8.0$\%                & $13.7$\%          \\
Whisper enc. + Gemma 7B                          & $8.6$\%                & $10.4$\%                           & $25.1$\%         & $11.7$\%               & $18.8$\%          \\
Whisper enc. + Llama 2 7B                        & $10.5$\%               & $12.8$\%                           & $22.5$\%         & $9.2$\%                & $14.8$\%          \\
Whisper enc. + Mistral 7B v0.1                   & $9.0$\%                & $10.2$\%                           & $23.7$\%         & $8.2$\%                & $14.5$\%          \\ \hline \hline
\end{tabularx}}
\caption{ASR evaluation results (WER)}
\label{tab:asr}
\end{table*}

\subsection{Offline ST Results}
\label{subsec:st}

\cref{tab:st_bleu,tab:st_comet} report the \textsc{BLEU} and \textsc{COMET}-family scores, respectively, on the four test sets. For evaluating with \textsc{BLEU}, we included both \texttt{docAsWhole} score, which concatenated all reference segments and candidate complete segments as two documents, and \texttt{mwerSegmenter} score, which resegments complete candidate segments according to reference segments to minimize WER. Similar to \cref{subsec:asr}, \texttt{mwerSegmenter} scores for IWSLT'21 test set could not be obtained, hence we did not include them.

Similarly, the models with Gemma 2 9B still have the best evaluation score among the four fine-tuned LLMs. In combination with the Whisper encoder, it even surpassed the performance of the cascaded system of Whisper + NLLB in most of the test sets and metrics.

\begin{table*}[h!]
\small
\centering
{\renewcommand{\arraystretch}{1.1}
\begin{tabular}{lcccc}
\hline
\multicolumn{1}{c}{\multirow{2}{*}{\textbf{Model}}} & \multicolumn{2}{c}{\textbf{MuST-C}}             & \multicolumn{2}{c}{\textbf{IWSLT}}  \\
\multicolumn{1}{c}{}                                & \textbf{tst-COMMON v2} & \textbf{tst-COMMON v3} & \textbf{tst2021} & \textbf{tst2022} \\ \hline \hline
Cascaded Whisper + NLLB                             & $39.84$ / $31.06$        & $40.30$ / $31.60$        & $\mathbf{43.84}$ / -       & $\mathbf{41.86}$ / $\mathbf{30.48}$  \\
SeamlessM4T                                         & $32.62$ / $22.98$        & $33.36$ / $23.59$        & $35.97$ / -       & $34.08$ / $22.68$  \\ \hline \hline
HuBERT + Gemma 2 9B                                 & $37.98$ / $28.15$        & $37.50$ / $27.59$        & $37.59$ / -       & $37.04$ / $25.86$  \\
HuBERT + Gemma 7B                                   & $36.20$ / $25.89$        & $36.24$ / $26.02$        & $33.00$ / -       & $34.27$ / $22.98$  \\
HuBERT + Llama 2 7B                                 & $36.52$ / $26.42$        & $35.93$ / $25.89$        & $35.66$ / -       & $35.13$ / $23.88$  \\
HuBERT + Mistral 7B v0.1                            & $36.91$ / $26.90$        & $36.94$ / $27.05$        & $36.29$ / -       & $36.09$ / $25.07$  \\ \hline
Whisper enc. + Gemma 2 9B                           & $\mathbf{41.33}$ / $\mathbf{31.98}$        & $\mathbf{41.16}$ / $\mathbf{31.72}$        & $40.76$ / -       & $39.64$ / $29.18$  \\
Whisper enc. + Gemma 7B                             & $38.62$ / $28.55$        & $38.81$ / $28.81$        & $37.02$ / -       & $37.58$ / $26.29$  \\
Whisper enc. + Llama 2 7B                           & $38.95$ / $29.17$        & $38.79$ / $28.94$        & $37.18$ / -       & $36.94$ / $26.18$  \\
Whisper enc. + Mistral 7B v0.1                      & $39.52$ / $30.03$        & $39.28$ / $29.59$        & $38.60$ / -       & $37.55$ / $26.64$  \\ \hline \hline
\end{tabular}}
\caption{Offline ST en2de \textsc{BLEU} results, with both \texttt{docAsWhole} and \texttt{mwerSegmenter} scores, respectively}
\label{tab:st_bleu}
\end{table*}

\begin{table*}[h!]
\small
\centering
{\renewcommand{\arraystretch}{1.1}
\begin{tabularx}{\linewidth}{p{0.3\linewidth}>{\centering}p{0.15\linewidth}>{\centering}p{0.15\linewidth}cc}
\hline
\multicolumn{1}{c}{\multirow{2}{*}{\textbf{Model}}} & \multicolumn{2}{c}{\textbf{MuST-C}}             & \multicolumn{2}{c}{\textbf{IWSLT}}  \\
\multicolumn{1}{c}{}                                & \textbf{tst-COMMON v2} & \textbf{tst-COMMON v3} & \textbf{tst2021} & \textbf{tst2022} \\ \hline \hline
Cascaded Whisper + NLLB                             & $83.00$ / $79.98$          & $82.49$ / $80.53$          & $64.47$ / $58.23$    & $65.32$ / $59.27$    \\
SeamlessM4T                                         & $76.72$ / $73.49$          & $76.42$ / $74.03$          & $59.63$ / $53.92$    & $60.34$ / $54.93$    \\ \hline \hline
HuBERT + Gemma 2 9B                                 & $80.98$ / $77.42$          & $80.17$ / $77.45$          & $67.63$ / $60.34$    & $67.11$ / $59.68$    \\
HuBERT + Gemma 7B                                   & $79.64$ / $75.52$          & $78.85$ / $75.53$          & $65.22$ / $57.51$    & $64.77$ / $57.23$    \\
HuBERT + Llama 2 7B                                 & $79.88$ / $76.30$          & $79.08$ / $76.32$          & $66.54$ / $59.27$    & $65.70$ / $58.70$    \\
HuBERT + Mistral 7B v0.1                            & $80.12$ / $76.92$          & $79.45$ / $76.92$          & $66.97$ / $59.73$    & $66.62$ / $59.85$    \\ \hline
Whisper enc. + Gemma 2 9B                           & $\mathbf{84.22}$ / $\mathbf{81.15}$          & $\mathbf{83.65}$ / $\mathbf{81.29}$          & $\mathbf{70.51}$ / $\mathbf{62.80}$    & $\mathbf{70.34}$ / $\mathbf{63.27}$    \\
Whisper enc. + Gemma 7B                             & $82.55$ / $79.69$          & $82.15$ / $79.88$          & $67.63$ / $60.06$    & $68.24$ / $60.91$    \\
Whisper enc. + Llama 2 7B                           & $82.84$ / $80.09$          & $82.14$ / $80.05$          & $68.82$ / $61.82$    & $68.64$ / $61.91$    \\
Whisper enc. + Mistral 7B v0.1                      & $83.13$ / $80.24$          & $82.43$ / $80.37$          & $69.73$ / $62.40$    & $68.86$ / $61.79$    \\ \hline \hline
\end{tabularx}}
\caption{Offline ST en2de $\text{COMET}^{\text{DA}}_{22}$ and $\text{COMET}^{\text{KIWI-DA}}_{22}$ results, respectively}
\label{tab:st_comet}
\end{table*}

\section{Future work}
\label{sec:future_work}
To date, we could only conduct experiments for the English-to-German direction; hence, in the future, we will expand our experiments to more language pairs and directions. In addition, we have some ideas to improve the pipeline:

\begin{itemize}
    \item Try replacing the CTC collapsing procedure with a length adapter of convolution layers for the HuBERT encoder. Try other modal adapter methods, like Q-Former.
    \item Experiment with smaller variants of the LLMs for faster training and inference, while retaining the robustness in translation, by distilling knowledge from fine-tuned systems.
    \item Integrate some reinforcement learning techniques into the pipeline for better performance.
\end{itemize}

\section{Conclusion}
\label{sec:concl}

In this paper, we leveraged pre-trained speech encoders and LLMs and connected them to become an end-to-end architecture for speech translation. The overall result is expected: for the English-to-German direction, even though our models performed better than the end-to-end SeamlessM4T model all of the time, there was still a gap compared to the performance of the cascaded Whisper + NLLB pipeline. It suggests that cascaded models are still the state-of-the-art approach in the speech translation task; this is also confirmed according to \citet{ahmad-etal-2024-findings}, in which all systems submitted to the Offline Track of IWSLT'24 were cascaded systems.

\section{Limitations}
The first problem we found was a limitation involving the sparse amount of parallel training data. This has been a notable issue for text translation, but for speech data, it is an even bigger concern, especially for low-resource languages. The two languages in our experiments, English and German, are considered high-resource languages, but the dataset only contains approximately 400 hours of audio.

Second, considering the size of the LLMs, our models were inferior regarding inference speed, compared to the two baselines. Our models also managed to surpass the performance of the cascaded system in the translation task; however, the differences were not too substantial. In addition, despite being a much smaller model, Whisper alone still excels at speech recognition. This raises a question: \textit{"Can end-to-end speech translation systems be smaller in size, while still keeping the robustness in translation, especially for the rising need to be used in mobile devices?"}

\section{Acknowledgment}
\label{sec:ack}
Computational resources were provided by the e-INFRA CZ project (ID:90254), supported by the Ministry of Education, Youth and Sports of the Czech Republic.

Nam Luu has been supported by the Erasmus Mundus program in Language and Communication Technologies (LCT).

Ondřej Bojar has received funding from the Project OP JAK Mezisektorová spolupráce Nr. CZ.02.01.01/00/23\_020/0008518 named ``Jazykověda, umělá inteligence a jazykové a řečové technologie: od výzkumu k aplikacím.''

\bibliography{custom}

\begin{thebibliography}{31}
\expandafter\ifx\csname natexlab\endcsname\relax\def\natexlab#1{#1}\fi

\bibitem[{Ahmad et~al.(2024)Ahmad, Anastasopoulos, Bojar, Borg, Carpuat, Cattoni, Cettolo, Chen, Dong, Federico, Haddow, Javorsk{\'y}, Krubi{\'n}ski, Kim~Lam, Ma, Mathur, Matusov, Maurya, McCrae, Murray, Nakamura, Negri, Niehues, Niu, Ojha, Ortega, Papi, Pol{\'a}k, Posp{\'\i}{\v{s}}il, Pecina, Salesky, Sethiya, Sarkar, Shi, Sikasote, Sperber, St{\"u}ker, Sudoh, Thompson, Waibel, Watanabe, Wilken, Zem{\'a}nek, and Zevallos}]{ahmad-etal-2024-findings}
Ibrahim~Said Ahmad, Antonios Anastasopoulos, Ond{\v{r}}ej Bojar, Claudia Borg, Marine Carpuat, Roldano Cattoni, Mauro Cettolo, William Chen, Qianqian Dong, Marcello Federico, Barry Haddow, D{\'a}vid Javorsk{\'y}, Mateusz Krubi{\'n}ski, Tsz Kim~Lam, Xutai Ma, Prashant Mathur, Evgeny Matusov, Chandresh Maurya, John McCrae, Kenton Murray, Satoshi Nakamura, Matteo Negri, Jan Niehues, Xing Niu, Atul~Kr. Ojha, John Ortega, Sara Papi, Peter Pol{\'a}k, Adam Posp{\'\i}{\v{s}}il, Pavel Pecina, Elizabeth Salesky, Nivedita Sethiya, Balaram Sarkar, Jiatong Shi, Claytone Sikasote, Matthias Sperber, Sebastian St{\"u}ker, Katsuhito Sudoh, Brian Thompson, Alex Waibel, Shinji Watanabe, Patrick Wilken, Petr Zem{\'a}nek, and Rodolfo Zevallos. 2024.
\newblock \href {https://aclanthology.org/2024.iwslt-1.1} {{FINDINGS} {OF} {THE} {IWSLT} 2024 {EVALUATION} {CAMPAIGN}}.
\newblock In \emph{Proceedings of the 21st International Conference on Spoken Language Translation (IWSLT 2024)}, pages 1--11, Bangkok, Thailand (in-person and online). Association for Computational Linguistics.

\bibitem[{Ansari et~al.(2021)Ansari, Bojar, Haddow, and Mahmoudi}]{ansari-etal-2021-sltev}
Ebrahim Ansari, Ond{\v{r}}ej Bojar, Barry Haddow, and Mohammad Mahmoudi. 2021.
\newblock \href {https://doi.org/10.18653/v1/2021.eacl-demos.9} {{{SLTEV}: Comprehensive Evaluation of Spoken Language Translation}}.
\newblock In \emph{Proceedings of the 16th Conference of the European Chapter of the Association for Computational Linguistics: System Demonstrations}, pages 71--79, Online. Association for Computational Linguistics.

\bibitem[{Bérard et~al.(2018)Bérard, Besacier, Kocabiyikoglu, and Pietquin}]{bérard2018endtoendautomaticspeechtranslation}
Alexandre Bérard, Laurent Besacier, Ali~Can Kocabiyikoglu, and Olivier Pietquin. 2018.
\newblock \href {http://arxiv.org/abs/1802.04200} {{End-to-End Automatic Speech Translation of Audiobooks}}.

\bibitem[{Cattoni et~al.(2021)Cattoni, {Di Gangi}, Bentivogli, Negri, and Turchi}]{CATTONI2021101155}
Roldano Cattoni, Mattia~Antonino {Di Gangi}, Luisa Bentivogli, Matteo Negri, and Marco Turchi. 2021.
\newblock \href {https://doi.org/https://doi.org/10.1016/j.csl.2020.101155} {{MuST-C: A multilingual corpus for end-to-end speech translation}}.
\newblock \emph{Computer Speech \& Language}, 66:101155.

\bibitem[{Communication et~al.(2023)Communication, Barrault, Chung, Meglioli, Dale, Dong, Duquenne, Elsahar, Gong, Heffernan, Hoffman, Klaiber, Li, Licht, Maillard, Rakotoarison, Sadagopan, Wenzek, Ye, Akula, Chen, Hachem, Ellis, Gonzalez, Haaheim, Hansanti, Howes, Huang, Hwang, Inaguma, Jain, Kalbassi, Kallet, Kulikov, Lam, Li, Ma, Mavlyutov, Peloquin, Ramadan, Ramakrishnan, Sun, Tran, Tran, Tufanov, Vogeti, Wood, Yang, Yu, Andrews, Balioglu, Costa-jussà, Celebi, Elbayad, Gao, Guzmán, Kao, Lee, Mourachko, Pino, Popuri, Ropers, Saleem, Schwenk, Tomasello, Wang, Wang, and Wang}]{communication2023seamlessm4tmassivelymultilingual}
Seamless Communication, Loïc Barrault, Yu-An Chung, Mariano~Cora Meglioli, David Dale, Ning Dong, Paul-Ambroise Duquenne, Hady Elsahar, Hongyu Gong, Kevin Heffernan, John Hoffman, Christopher Klaiber, Pengwei Li, Daniel Licht, Jean Maillard, Alice Rakotoarison, Kaushik~Ram Sadagopan, Guillaume Wenzek, Ethan Ye, Bapi Akula, Peng-Jen Chen, Naji~El Hachem, Brian Ellis, Gabriel~Mejia Gonzalez, Justin Haaheim, Prangthip Hansanti, Russ Howes, Bernie Huang, Min-Jae Hwang, Hirofumi Inaguma, Somya Jain, Elahe Kalbassi, Amanda Kallet, Ilia Kulikov, Janice Lam, Daniel Li, Xutai Ma, Ruslan Mavlyutov, Benjamin Peloquin, Mohamed Ramadan, Abinesh Ramakrishnan, Anna Sun, Kevin Tran, Tuan Tran, Igor Tufanov, Vish Vogeti, Carleigh Wood, Yilin Yang, Bokai Yu, Pierre Andrews, Can Balioglu, Marta~R. Costa-jussà, Onur Celebi, Maha Elbayad, Cynthia Gao, Francisco Guzmán, Justine Kao, Ann Lee, Alexandre Mourachko, Juan Pino, Sravya Popuri, Christophe Ropers, Safiyyah Saleem, Holger Schwenk, Paden Tomasello, Changhan Wang, Jeff
  Wang, and Skyler Wang. 2023.
\newblock \href {http://arxiv.org/abs/2308.11596} {{SeamlessM4T: Massively Multilingual \& Multimodal Machine Translation}}.

\bibitem[{Dettmers et~al.(2023)Dettmers, Pagnoni, Holtzman, and Zettlemoyer}]{dettmers2023qloraefficientfinetuningquantized}
Tim Dettmers, Artidoro Pagnoni, Ari Holtzman, and Luke Zettlemoyer. 2023.
\newblock \href {http://arxiv.org/abs/2305.14314} {{QLoRA: Efficient Finetuning of Quantized LLMs}}.

\bibitem[{Gaido et~al.(2021)Gaido, Cettolo, Negri, and Turchi}]{gaido2021ctcbasedcompressiondirectspeech}
Marco Gaido, Mauro Cettolo, Matteo Negri, and Marco Turchi. 2021.
\newblock \href {http://arxiv.org/abs/2102.01578} {{CTC-based Compression for Direct Speech Translation}}.

\bibitem[{Gaido et~al.(2020)Gaido, Gangi, Negri, and Turchi}]{gaido2020endtoendspeechtranslationknowledgedistillation}
Marco Gaido, Mattia Antonino~Di Gangi, Matteo Negri, and Marco Turchi. 2020.
\newblock \href {http://arxiv.org/abs/2006.02965} {{End-to-End Speech-Translation with Knowledge Distillation: FBK@IWSLT2020}}.

\bibitem[{Gao et~al.(2023)Gao, Han, Zhang, Lin, Geng, Zhou, Zhang, Lu, He, Yue, Li, and Qiao}]{gao2023llamaadapterv2parameterefficientvisual}
Peng Gao, Jiaming Han, Renrui Zhang, Ziyi Lin, Shijie Geng, Aojun Zhou, Wei Zhang, Pan Lu, Conghui He, Xiangyu Yue, Hongsheng Li, and Yu~Qiao. 2023.
\newblock \href {http://arxiv.org/abs/2304.15010} {{LLaMA-Adapter V2: Parameter-Efficient Visual Instruction Model}}.

\bibitem[{Hsu et~al.(2021)Hsu, Bolte, Tsai, Lakhotia, Salakhutdinov, and Mohamed}]{hsu2021hubertselfsupervisedspeechrepresentation}
Wei-Ning Hsu, Benjamin Bolte, Yao-Hung~Hubert Tsai, Kushal Lakhotia, Ruslan Salakhutdinov, and Abdelrahman Mohamed. 2021.
\newblock \href {http://arxiv.org/abs/2106.07447} {{HuBERT: Self-Supervised Speech Representation Learning by Masked Prediction of Hidden Units}}.

\bibitem[{Kahn et~al.(2020)Kahn, Riviere, Zheng, Kharitonov, Xu, Mazare, Karadayi, Liptchinsky, Collobert, Fuegen, Likhomanenko, Synnaeve, Joulin, Mohamed, and Dupoux}]{Kahn_2020}
J.~Kahn, M.~Riviere, W.~Zheng, E.~Kharitonov, Q.~Xu, P.E. Mazare, J.~Karadayi, V.~Liptchinsky, R.~Collobert, C.~Fuegen, T.~Likhomanenko, G.~Synnaeve, A.~Joulin, A.~Mohamed, and E.~Dupoux. 2020.
\newblock \href {https://doi.org/10.1109/icassp40776.2020.9052942} {{Libri-Light: A Benchmark for ASR with Limited or No Supervision}}.
\newblock In \emph{ICASSP 2020 - 2020 IEEE International Conference on Acoustics, Speech and Signal Processing (ICASSP)}. IEEE.

\bibitem[{Li et~al.(2023{\natexlab{a}})Li, Li, Savarese, and Hoi}]{li2023blip2bootstrappinglanguageimagepretraining}
Junnan Li, Dongxu Li, Silvio Savarese, and Steven Hoi. 2023{\natexlab{a}}.
\newblock \href {http://arxiv.org/abs/2301.12597} {{BLIP-2: Bootstrapping Language-Image Pre-training with Frozen Image Encoders and Large Language Models}}.

\bibitem[{Li et~al.(2023{\natexlab{b}})Li, Wu, Li, and Liu}]{li2023promptinglargelanguagemodels}
Yuang Li, Yu~Wu, Jinyu Li, and Shujie Liu. 2023{\natexlab{b}}.
\newblock \href {http://arxiv.org/abs/2306.16007} {{Prompting Large Language Models for Zero-Shot Domain Adaptation in Speech Recognition}}.

\bibitem[{Liu et~al.(2023)Liu, Li, Wu, and Lee}]{liu2023visualinstructiontuning}
Haotian Liu, Chunyuan Li, Qingyang Wu, and Yong~Jae Lee. 2023.
\newblock \href {http://arxiv.org/abs/2304.08485} {{Visual Instruction Tuning}}.

\bibitem[{Liu et~al.(2019)Liu, Xiong, He, Zhang, Wu, Wang, and Zong}]{liu2019endtoendspeechtranslationknowledge}
Yuchen Liu, Hao Xiong, Zhongjun He, Jiajun Zhang, Hua Wu, Haifeng Wang, and Chengqing Zong. 2019.
\newblock \href {http://arxiv.org/abs/1904.08075} {{End-to-End Speech Translation with Knowledge Distillation}}.

\bibitem[{Loshchilov and Hutter(2017)}]{loshchilov2017sgdrstochasticgradientdescent}
Ilya Loshchilov and Frank Hutter. 2017.
\newblock \href {http://arxiv.org/abs/1608.03983} {{SGDR: Stochastic Gradient Descent with Warm Restarts}}.

\bibitem[{Loshchilov and Hutter(2019)}]{loshchilov_decoupled_adamw_2019}
Ilya Loshchilov and Frank Hutter. 2019.
\newblock \href {https://doi.org/10.48550/arXiv.1711.05101} {Decoupled {Weight} {Decay} {Regularization}}.
\newblock ArXiv:1711.05101 [cs, math].

\bibitem[{Minaee et~al.(2024)Minaee, Mikolov, Nikzad, Chenaghlu, Socher, Amatriain, and Gao}]{minaee2024largelanguagemodelssurvey}
Shervin Minaee, Tomas Mikolov, Narjes Nikzad, Meysam Chenaghlu, Richard Socher, Xavier Amatriain, and Jianfeng Gao. 2024.
\newblock \href {http://arxiv.org/abs/2402.06196} {{Large Language Models: A Survey}}.

\bibitem[{Naveed et~al.(2024)Naveed, Khan, Qiu, Saqib, Anwar, Usman, Akhtar, Barnes, and Mian}]{naveed2024comprehensiveoverviewlargelanguage}
Humza Naveed, Asad~Ullah Khan, Shi Qiu, Muhammad Saqib, Saeed Anwar, Muhammad Usman, Naveed Akhtar, Nick Barnes, and Ajmal Mian. 2024.
\newblock \href {http://arxiv.org/abs/2307.06435} {{A Comprehensive Overview of Large Language Models}}.

\bibitem[{Panayotov et~al.(2015{\natexlab{a}})Panayotov, Chen, Povey, and Khudanpur}]{librispeech7178964}
Vassil Panayotov, Guoguo Chen, Daniel Povey, and Sanjeev Khudanpur. 2015{\natexlab{a}}.
\newblock \href {https://doi.org/10.1109/ICASSP.2015.7178964} {{LibriSpeech: An ASR corpus based on public domain audio books}}.
\newblock In \emph{2015 IEEE International Conference on Acoustics, Speech and Signal Processing (ICASSP)}, pages 5206--5210.

\bibitem[{Panayotov et~al.(2015{\natexlab{b}})Panayotov, Chen, Povey, and Khudanpur}]{7178964}
Vassil Panayotov, Guoguo Chen, Daniel Povey, and Sanjeev Khudanpur. 2015{\natexlab{b}}.
\newblock \href {https://doi.org/10.1109/ICASSP.2015.7178964} {Librispeech: An asr corpus based on public domain audio books}.
\newblock In \emph{2015 IEEE International Conference on Acoustics, Speech and Signal Processing (ICASSP)}, pages 5206--5210.

\bibitem[{Papineni et~al.(2002)Papineni, Roukos, Ward, and Zhu}]{bleu}
Kishore Papineni, Salim Roukos, Todd Ward, and Wei-Jing Zhu. 2002.
\newblock \href {https://doi.org/10.3115/1073083.1073135} {{BLEU: a method for automatic evaluation of machine translation}}.
\newblock In \emph{Proceedings of the 40th Annual Meeting on Association for Computational Linguistics}, ACL '02, page 311–318, USA. Association for Computational Linguistics.

\bibitem[{Post(2018)}]{post2018clarityreportingbleuscores}
Matt Post. 2018.
\newblock \href {http://arxiv.org/abs/1804.08771} {{A Call for Clarity in Reporting BLEU Scores}}.

\bibitem[{Radford et~al.(2022)Radford, Kim, Xu, Brockman, McLeavey, and Sutskever}]{radford2022robustspeechrecognitionlargescale}
Alec Radford, Jong~Wook Kim, Tao Xu, Greg Brockman, Christine McLeavey, and Ilya Sutskever. 2022.
\newblock \href {http://arxiv.org/abs/2212.04356} {{Robust Speech Recognition via Large-Scale Weak Supervision}}.

\bibitem[{Rei et~al.(2022{\natexlab{a}})Rei, C.~de Souza, Alves, Zerva, Farinha, Glushkova, Lavie, Coheur, and Martins}]{rei-etal-2022-comet}
Ricardo Rei, Jos{\'e}~G. C.~de Souza, Duarte Alves, Chrysoula Zerva, Ana~C Farinha, Taisiya Glushkova, Alon Lavie, Luisa Coheur, and Andr{\'e} F.~T. Martins. 2022{\natexlab{a}}.
\newblock \href {https://aclanthology.org/2022.wmt-1.52} {{{COMET}-22: Unbabel-{IST} 2022 Submission for the Metrics Shared Task}}.
\newblock In \emph{Proceedings of the Seventh Conference on Machine Translation (WMT)}, pages 578--585, Abu Dhabi, United Arab Emirates (Hybrid). Association for Computational Linguistics.

\bibitem[{Rei et~al.(2022{\natexlab{b}})Rei, Treviso, Guerreiro, Zerva, Farinha, Maroti, de~Souza, Glushkova, Alves, Lavie, Coheur, and Martins}]{rei2022cometkiwiistunbabel2022submission}
Ricardo Rei, Marcos Treviso, Nuno~M. Guerreiro, Chrysoula Zerva, Ana~C. Farinha, Christine Maroti, José G.~C. de~Souza, Taisiya Glushkova, Duarte~M. Alves, Alon Lavie, Luisa Coheur, and André F.~T. Martins. 2022{\natexlab{b}}.
\newblock \href {http://arxiv.org/abs/2209.06243} {{CometKiwi: IST-Unbabel 2022 Submission for the Quality Estimation Shared Task}}.

\bibitem[{Team et~al.(2022)Team, Costa-jussà, Cross, Çelebi, Elbayad, Heafield, Heffernan, Kalbassi, Lam, Licht, Maillard, Sun, Wang, Wenzek, Youngblood, Akula, Barrault, Gonzalez, Hansanti, Hoffman, Jarrett, Sadagopan, Rowe, Spruit, Tran, Andrews, Ayan, Bhosale, Edunov, Fan, Gao, Goswami, Guzmán, Koehn, Mourachko, Ropers, Saleem, Schwenk, and Wang}]{nllbteam2022languageleftbehindscaling}
NLLB Team, Marta~R. Costa-jussà, James Cross, Onur Çelebi, Maha Elbayad, Kenneth Heafield, Kevin Heffernan, Elahe Kalbassi, Janice Lam, Daniel Licht, Jean Maillard, Anna Sun, Skyler Wang, Guillaume Wenzek, Al~Youngblood, Bapi Akula, Loic Barrault, Gabriel~Mejia Gonzalez, Prangthip Hansanti, John Hoffman, Semarley Jarrett, Kaushik~Ram Sadagopan, Dirk Rowe, Shannon Spruit, Chau Tran, Pierre Andrews, Necip~Fazil Ayan, Shruti Bhosale, Sergey Edunov, Angela Fan, Cynthia Gao, Vedanuj Goswami, Francisco Guzmán, Philipp Koehn, Alexandre Mourachko, Christophe Ropers, Safiyyah Saleem, Holger Schwenk, and Jeff Wang. 2022.
\newblock \href {http://arxiv.org/abs/2207.04672} {{No Language Left Behind: Scaling Human-Centered Machine Translation}}.

\bibitem[{Wu et~al.(2023)Wu, Gaur, Chen, Zhou, Zhu, Wang, Li, Liu, Ren, Liu, and Wu}]{wu2023decoderonlyarchitecturespeechtotextlarge}
Jian Wu, Yashesh Gaur, Zhuo Chen, Long Zhou, Yimeng Zhu, Tianrui Wang, Jinyu Li, Shujie Liu, Bo~Ren, Linquan Liu, and Yu~Wu. 2023.
\newblock \href {http://arxiv.org/abs/2307.03917} {{On decoder-only architecture for speech-to-text and large language model integration}}.

\bibitem[{Zhang et~al.(2023)Zhang, Li, Zhang, Zhan, Wang, Zhou, and Qiu}]{zhang2023speechgptempoweringlargelanguage}
Dong Zhang, Shimin Li, Xin Zhang, Jun Zhan, Pengyu Wang, Yaqian Zhou, and Xipeng Qiu. 2023.
\newblock \href {http://arxiv.org/abs/2305.11000} {{SpeechGPT: Empowering Large Language Models with Intrinsic Cross-Modal Conversational Abilities}}.

\bibitem[{Zhang et~al.(2024)Zhang, Dong, Li, Zhang, Sun, Wang, Li, Hu, Zhang, Wu, and Wang}]{zhang2024instructiontuninglargelanguage}
Shengyu Zhang, Linfeng Dong, Xiaoya Li, Sen Zhang, Xiaofei Sun, Shuhe Wang, Jiwei Li, Runyi Hu, Tianwei Zhang, Fei Wu, and Guoyin Wang. 2024.
\newblock \href {http://arxiv.org/abs/2308.10792} {{Instruction Tuning for Large Language Models: A Survey}}.

\bibitem[{Zhao et~al.(2023)Zhao, Zhou, Li, Tang, Wang, Hou, Min, Zhang, Zhang, Dong, Du, Yang, Chen, Chen, Jiang, Ren, Li, Tang, Liu, Liu, Nie, and Wen}]{zhao2023surveylargelanguagemodels}
Wayne~Xin Zhao, Kun Zhou, Junyi Li, Tianyi Tang, Xiaolei Wang, Yupeng Hou, Yingqian Min, Beichen Zhang, Junjie Zhang, Zican Dong, Yifan Du, Chen Yang, Yushuo Chen, Zhipeng Chen, Jinhao Jiang, Ruiyang Ren, Yifan Li, Xinyu Tang, Zikang Liu, Peiyu Liu, Jian-Yun Nie, and Ji-Rong Wen. 2023.
\newblock \href {http://arxiv.org/abs/2303.18223} {{A Survey of Large Language Models}}.

\end{thebibliography}

\appendix
\section{Training and Inference Details}
All models were fine-tuned using 4-bit QLoRA \citep{dettmers2023qloraefficientfinetuningquantized} adapters in \texttt{bfloat16} precision, with the following LoRA parameters: rank of $r=8$, alpha of $\alpha=8$. For the models with HuBERT as the encoder, because of the manual CTC collapsing procedure, we could only process one example at a time, hence the batch size was set to $1$; while for those with Whisper, the batch size was set to $2$. Other training hyperparameters included the learning rate of $1e-4$ with $10$ warmup steps, and an AdamW optimizer \cite{loshchilov_decoupled_adamw_2019} with a cosine scheduler \cite{loshchilov2017sgdrstochasticgradientdescent}. All HuBERT-encoder models were trained for $500,000$ steps, while Whisper-encoder models were trained for $100,000$ steps. 

\begin{figure}[h]
  \centering
  \begin{subfigure}[h]{\linewidth}
    \centering
    \includegraphics[width=\linewidth]{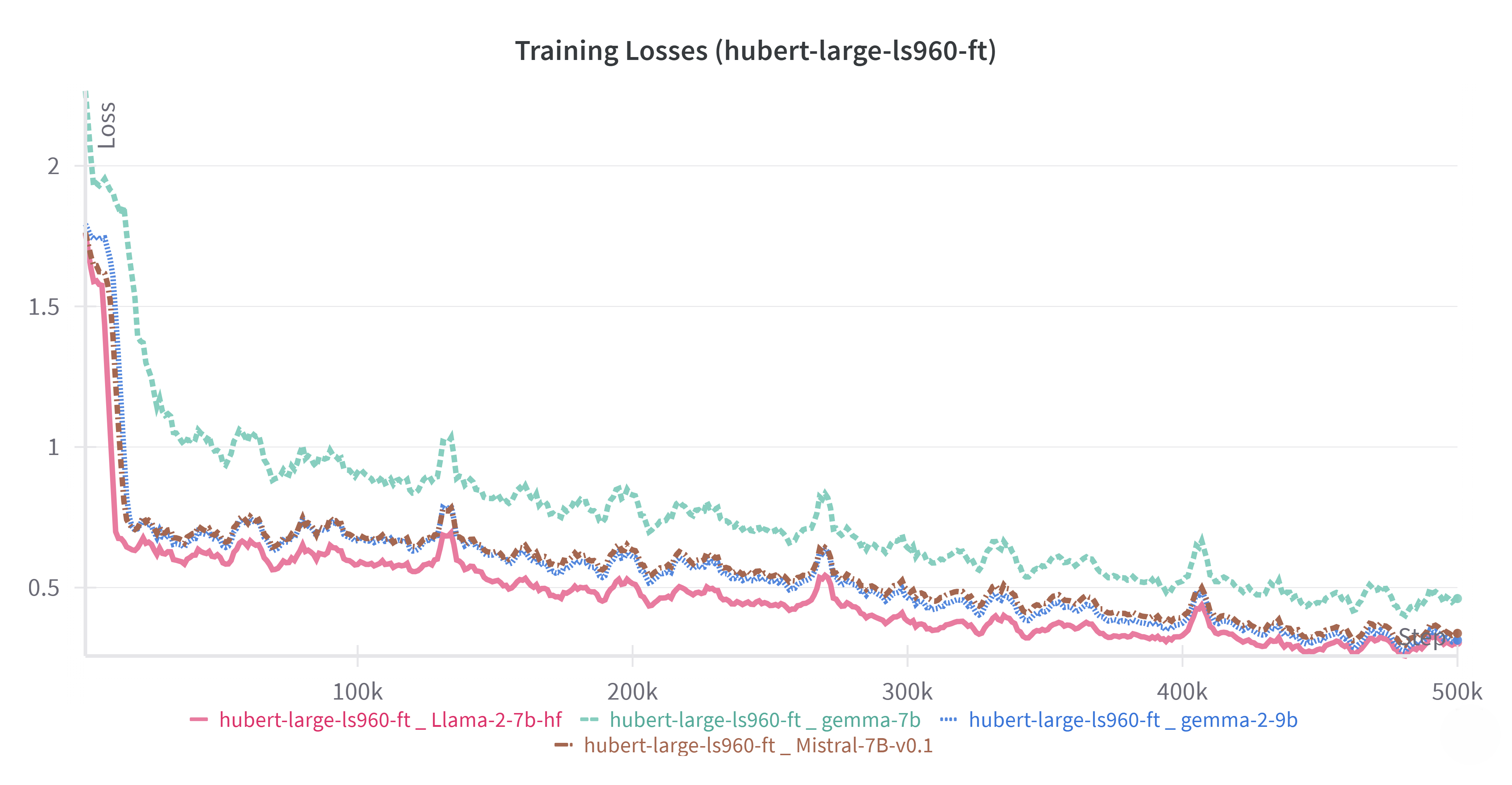}
    \caption{With HuBERT encoder}
    \label{fig:hubert_loss}
 \end{subfigure}
 \hfill
 \begin{subfigure}[h]{\linewidth}
    \centering
    \includegraphics[width=\linewidth]{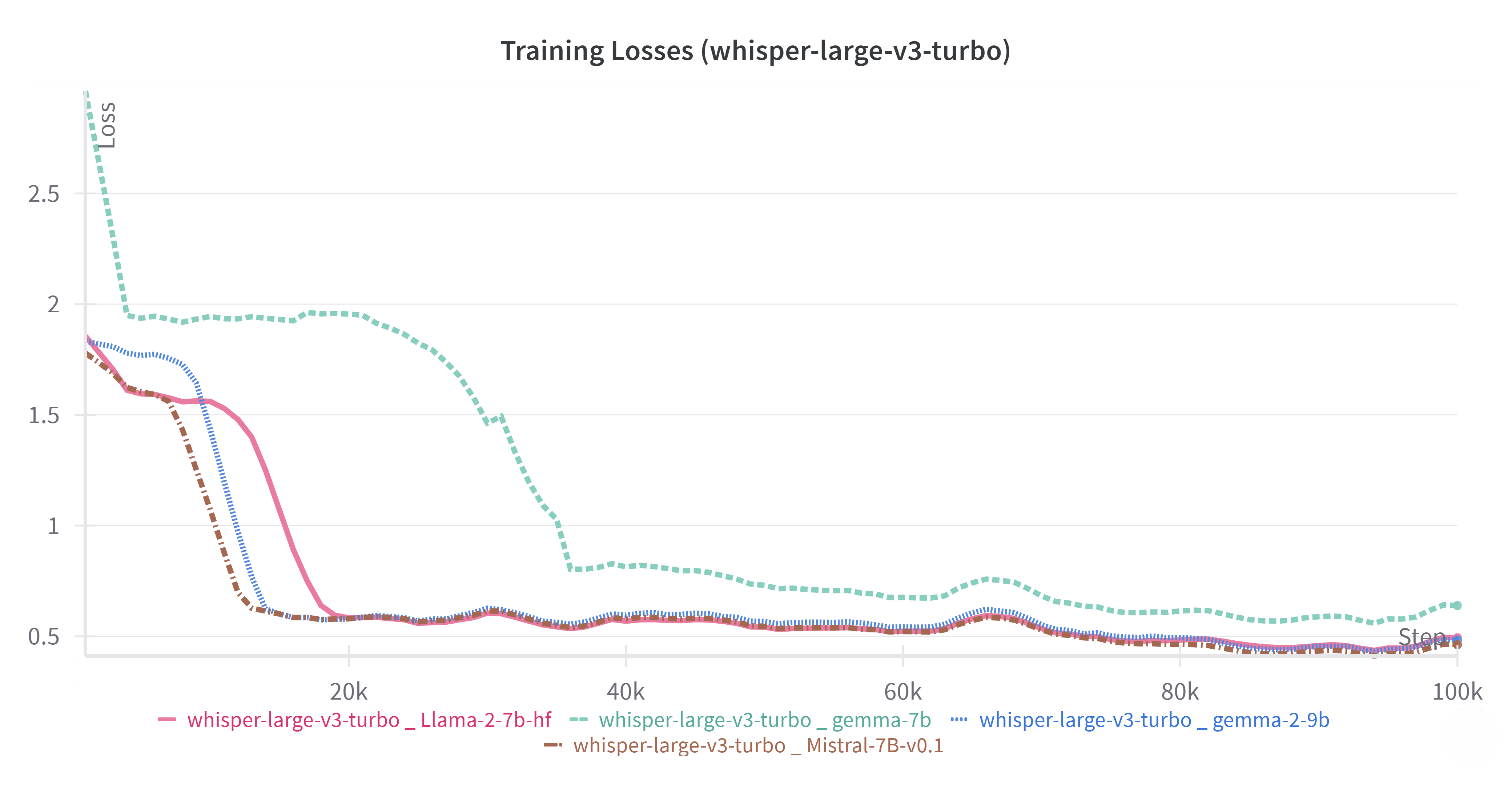}
    \caption{With Whisper encoder}
    \label{fig:whisper_loss}
 \end{subfigure}
  \caption{Training loss of models}
\end{figure}

During training, we added three new tokens to feed into the LLMs, namely ``\texttt{<>audio<>}'', ``\texttt{<>transcript<>}'', and ``\texttt{<>translation<>}'', which acted as separators between the extracted audio features, the ASR transcript, and the corresponding translation, respectively. For each sample, the training data is formatted as follows: ``\texttt{<bos> <>audio<> \{audio features\} <>transcript<> \{transcript\} <>translation<> \{translation\} <eos>}''. The cross-entropy loss was computed only for the tokens following ``\texttt{<>transcript<>}''. Each model's training loss details are illustrated in \cref{fig:hubert_loss,fig:whisper_loss}.

During inference, for each audio data, the LLMs were prompted using the following format: ``\texttt{<bos> <>audio<> \{audio features\} <>transcript<>}'', then generated the transcript and the corresponding translated text in an auto-regressive manner. 
We performed inference using the beam search algorithm, with a beam size of 2 for all models. All evaluation results, 
are described in \cref{subsec:asr,subsec:st}.




\end{document}